\pdfoutput=1

\documentclass[11pt]{article}

\usepackage{EMNLP2022}

\usepackage{times}
\usepackage{latexsym}

\usepackage[T1]{fontenc}

\usepackage[utf8]{inputenc}

\usepackage{microtype}

\usepackage{inconsolata}

\usepackage{times}
\usepackage{latexsym}

\usepackage{microtype}

\usepackage{amssymb}
\usepackage{bm}
\usepackage{times}

\usepackage{graphicx}
\usepackage{multirow}
\usepackage{enumerate}
\usepackage{booktabs}
\usepackage{enumitem}
\usepackage{amsmath}
\usepackage{hyperref}

\title{UniGeo: Unifying Geometry Logical Reasoning via  Reformulating Mathematical Expression}

\author{Jiaqi Chen$^{1,3}$,~~ Tong Li$^{2}$,~~ Jinghui Qin$^{4}$, ~~ Pan Lu$^{5}$, \\ 
~~ {\bf Liang Lin}$^{1}${\bf ,} ~~{\bf Chongyu Chen}$^{6}${\bf ,} ~~ {\bf Xiaodan Liang}$^{1,2}$\thanks{~~Corresponding author.}~\\
{$^1$Sun Yat-sen University, $^2$Shenzhen Campus of Sun Yat-sen University, } \\
{ $^3$The University of Hong Kong, $^4$Guangdong University of Technology, }\\ {$^5$University of California, Los Angeles, $^6$DarkMatter AI Research}\\
}

\begin{document}
\maketitle

\begin{abstract}
Geometry problem solving is a well-recognized testbed for evaluating the high-level multi-modal reasoning capability of deep models. In most existing works, two main geometry problems: calculation and proving, are usually treated as two specific tasks, hindering a deep model to unify its reasoning capability on multiple math tasks. However, in essence, these two tasks have similar problem representations and overlapped math knowledge which can improve the understanding and reasoning ability of a deep model on both two tasks.
Therefore, we construct a large-scale \textbf{Uni}fied \textbf{Geo}metry problem benchmark, \textbf{UniGeo}, which contains 4,998 calculation problems and 9,543 proving problems.
Each proving problem is annotated with a multi-step proof with reasons and mathematical expressions. The proof can be easily reformulated as a proving sequence that shares the same formats with the annotated program sequence for calculation problems.
Naturally, we also present a unified multi-task \textbf{Geo}metric Trans\textbf{former} framework, \textbf{Geoformer}, to tackle calculation and proving problems simultaneously in the form of sequence generation, which finally shows the reasoning ability can be improved on both two tasks by unifying formulation. Furthermore, we propose a Mathematical Expression Pretraining (MEP) method that aims to predict the mathematical expressions in the problem solution, thus improving the Geoformer model.
Experiments on the UniGeo demonstrate that our proposed Geoformer obtains state-of-the-art performance by outperforming task-specific model NGS with over 5.6\% and 3.2\% accuracies on calculation and proving problems, respectively.\footnote{Data and code: \href{https://github.com/chen-judge/UniGeo}{https://github.com/chen-judge/UniGeo} \href{https://github.com/chen-judge/UniGeo-MindSpore}{https://github.com/chen-judge/UniGeo-MindSpore}}

\end{abstract}

\begin{figure}[t]
\begin{center}
 \includegraphics[width=1.0\columnwidth]{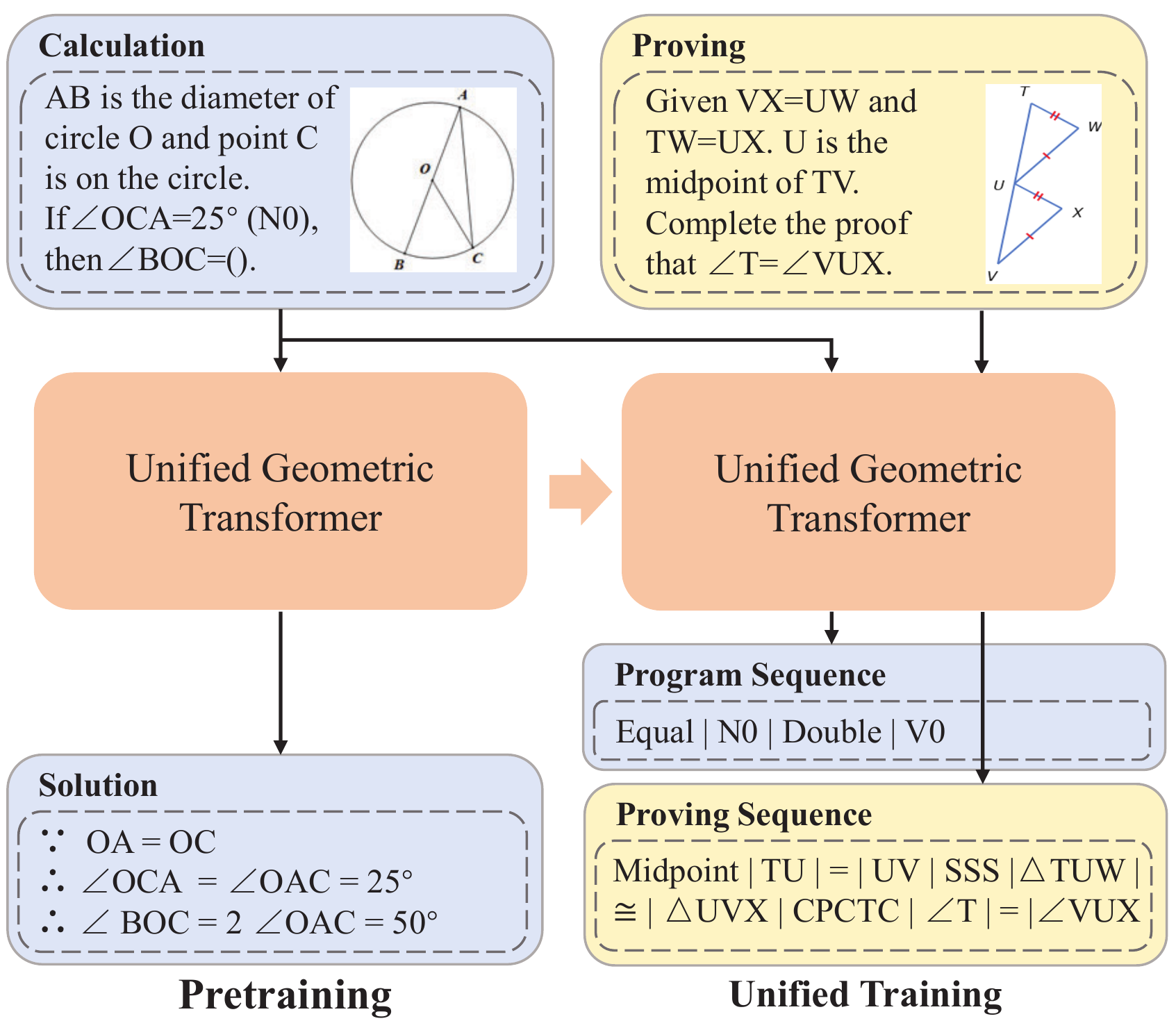}
 \vspace{-8mm}
\end{center}
  \caption{
  The pipeline of pretraining and unified training of our proposed Geoformer. We pretrain the model by predicting the mathematical expression extracted from the solution of calculation problems. After that, we consider the calculation and proving as downstream tasks, and feed both types of data to Geoformer for unified training. 
  }
\vspace{-2mm}
\label{fig:pipeline}
\end{figure}

\begin{figure*}[t]
\begin{center}
 \includegraphics[width=0.98\textwidth]{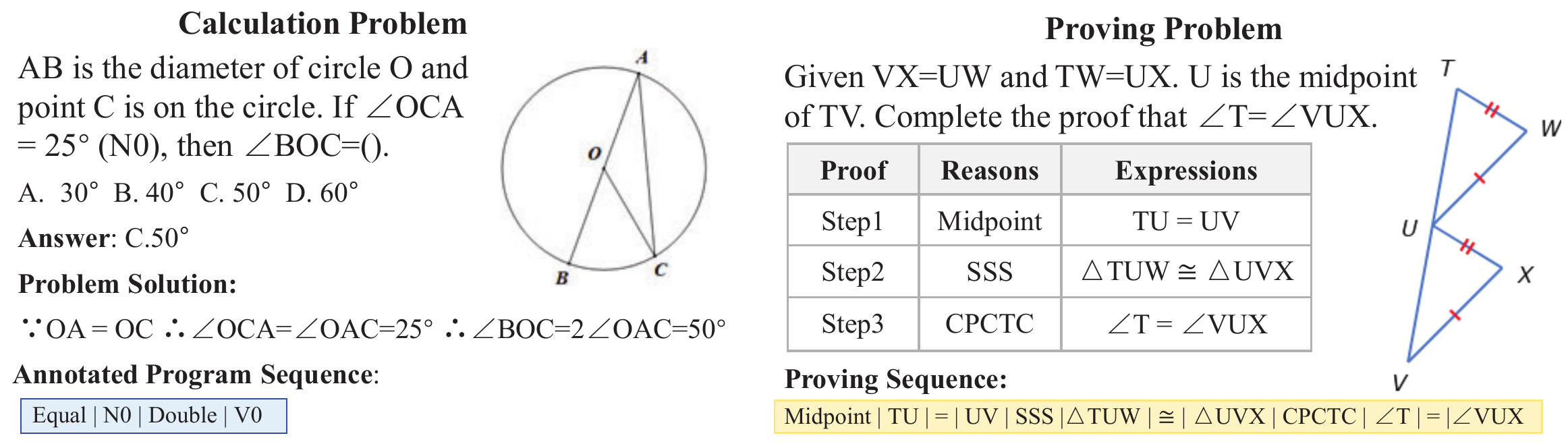}
 \vspace{-3mm}
\end{center}
  \caption{
  We unify geometry logical reasoning in the proposed UniGeo dataset. Except for the calculation problem provided in the GeoQA benchmark \cite{chen2021geoqa}, we collect some proving problems (right) which contain clear mathematical expressions and corresponding reasons that can be reformulated as proving sequence to unify with the program sequence in the calculation problems.
  }
\label{fig:introduce}
\end{figure*}

\section{Introduction}

Achieving logical reasoning abilities is still challenging for neural networks, especially in some mathematical reasoning tasks, such as math word problems (MWP) \cite{tsrmd,graph2tree,sau-solver,NS-Solver,yang-etal-2022-unbiased,yang2022logicsolver,mishra2022lila,mishra2022numglue,lu2022dynamic}, mathematical theorem proving \cite{li2020isarstep,welleck2021naturalproofs}, etc. 
Recently, geometry problem solving \cite{sachan2020discourse,chen2021geoqa, lu2021inter, ijcai2022p228} has also attracted much attention in the NLP community, which requires comprehensive reasoning capabilities in parsing multimodal information and utilizing mathematical knowledge.
Specifically, geometry problem solving mainly contains two categories: calculation and proving.
For calculation problems, both recent GeoQA \cite{chen2021geoqa} and Inter-GPS \cite{lu2021inter} propose multiple-choice geometry problem benchmarks annotated with specific symbolic programs or logic forms, which inspire the neural networks' potential ability to give an interpretable prediction. 
On the subject of geometry proving, the existing work \cite{chou1996automated,chou2000deductive,gan2019automatically} mainly relies on well-designed proving systems and forward chaining search methods rather than neural-based models.
Therefore, there is still a huge gap between the works on these two types of geometry problems, which are usually considered as two areas.

Recently, much work \cite{raffel2020exploring,cho2021unifying,lu2021iconqa,li2022blip,alayrac2022flamingo} has presented unified models for various vision-language reasoning and generation tasks since the underlying visual/linguistic understanding and reasoning abilities are largely common.
Inspired by the mainstream progress, we suppose that a unified model for geometry problem solving is also necessary.
To begin, calculation and proving tasks share some fundamental skills and knowledge in geometric reasoning. Therefore, it is desirable to explore the general understanding and reasoning ability of the unified neural network in the math domain.
Besides, the unified model doesn't need auxiliary models to determine whether the problem is a calculation or proving problem and further select task-specific models, where cumulative errors can be introduced.

To this end, a framework addressing geometry problems uniformly at both the data level and the model level is valuable and expected. However, the existing proving data is small-scale and annotated in an incompatible format. To achieve our goal, we collect lots of geometry proving data from an online education website and build a single multi-task benchmark, \textbf{UniGeo}, in which the provided proof can be reformulated as a causal proving sequence so that the calculation and proving problems are unified in data format, as shown in Figure \ref{fig:introduce}.
Our UniGeo contains 4,998 calculation problems and 9,543 proving problems, which can verify the high-level geometry logical reasoning capabilities in neural models.

Taking advantage of the unified formulation of two geometry tasks, we further propose a novel unified \textbf{geo}metric trans\textbf{former} (\textbf{Geoformer}) which is able to handle geometry calculation and proof reasoning simultaneously and outperforms the task-specialized models on both tasks.
To learn an efficient Geoformer for unifying geometry logical reasoning, we also propose a mathematical reasoning pre-training method named Mathematical Expression Pretraining (MEP), which is based on the problem solution, since the solution prediction can serve as a universal task for all math problems. Specifically, we extract the mathematical expressions and remove the redundant text description in the solution for MEP. These expressions are rich in implicit math knowledge and can also be formulated as the solution sequence target. We further fine-tune the unified Geoformer to predict program/proving sequences for calculation and proving problems simultaneously.
The pipeline of pretraining and unified training is demonstrated in Figure \ref{fig:pipeline}.
Experiments on the UniGeo benchmark show that our proposed Geoformer achieves state-of-the-art performance, getting 5.6\% and 3.2\% accuracy improvements on calculation and proving problems, respectively, compared to the task-specific model NGS \cite{chen2021geoqa}.

Our contributions can be summarized as follows:
\begin{itemize} 
\setlength{\itemsep}{0pt}
\setlength{\parsep}{0pt}
\setlength{\parskip}{0pt}
    \item We construct a unified geometry reasoning benchmark, named UniGeo, which contains both calculation and proving problems. 
    \item The proving problems in UniGeo are annotated with proof steps, in which the mathematical expressions can be reformulated as proving sequences to match the program sequences in calculation problems.
    \item We propose a unified geometric transformer framework, which is pretrained by predicting mathematical expressions in the solution and then fine-tuned on calculation and proving problems simultaneously.
\end{itemize}

\section{Related Work}
\paragraph{Geometry Problem Solving}

Several geometry datasets \cite{seo2014diagram,seo2015solving,sachan2017textbooks,alvin2017synthesis,sachan2017learning} have been constructed to facilitate the development of geometry problem solving. However, these previous geometry datasets are either not publicly available or built up with small sizes, which limit the development of relevant research. Besides, the latest datasets \cite{lu2021inter, chen2021geoqa, cao2022augmented} only focus on the arithmetic calculation skill for geometry problem solving and fail to take into account comprehensive geometry reasoning abilities like logical proving. 
For instance, GeoQA \cite{chen2021geoqa} provides 4,998 calculation problems annotated with a symbolic program sequence that corresponds to the problem solution.
Instead, we propose a new large-scale geometry dataset, which covers a wide range of sub-tasks and reasoning skills including calculation and proving.
To the best of our knowledge, we are the first work to collect so many geometry proving problems for training the neural network and provide detailed sequence annotations corresponding to the proofs which can be unified with calculation problems and facilitate model learning.

\paragraph{Geometry Theorem Proving} Theorem proving in the geometry domain \cite{gelernter1960empirical,chou1996automated,chou2000deductive,jgex2011,yu2019framework,gan2019automatically} is a long-standing artificial intelligence task. For example, \cite{chou1996automated} developed an initial automated geometry theorem proving system by designing a set of full-angle-based rules. Similarly, the expert system JGEX \cite{jgex2011} is proposed to prove full-angle geometry problems with a well-defined deductive database. 
More recently, some pioneering efforts \cite{li2020isarstep,tafjord2020proofwriter,welleck2021naturalproofs} have been attempted to learn automatic proofing systems from large-scale natural language corpus or mathematical propositions. 
However, how to achieve an automatic neural-based prover in the geometry domain is still less studied. Therefore, we propose a unified Geoformer that can generate proof given a geometry diagram and statements from scratch.

\section{Unifying Geometry Reasoning}

In this work, we aim to unify geometry logical reasoning for both calculation and proving problems.
To this end, we first construct a geometry proving dataset that requires multiple reasoning abilities while solving the problems. Furthermore, we reformulate the proof as sequence form which is consistent with the program sequence in the calculation problems of the current GeoQA \cite{chen2021geoqa}.

\begin{figure*}[t]
\begin{center}
 \includegraphics[width=0.95\textwidth]{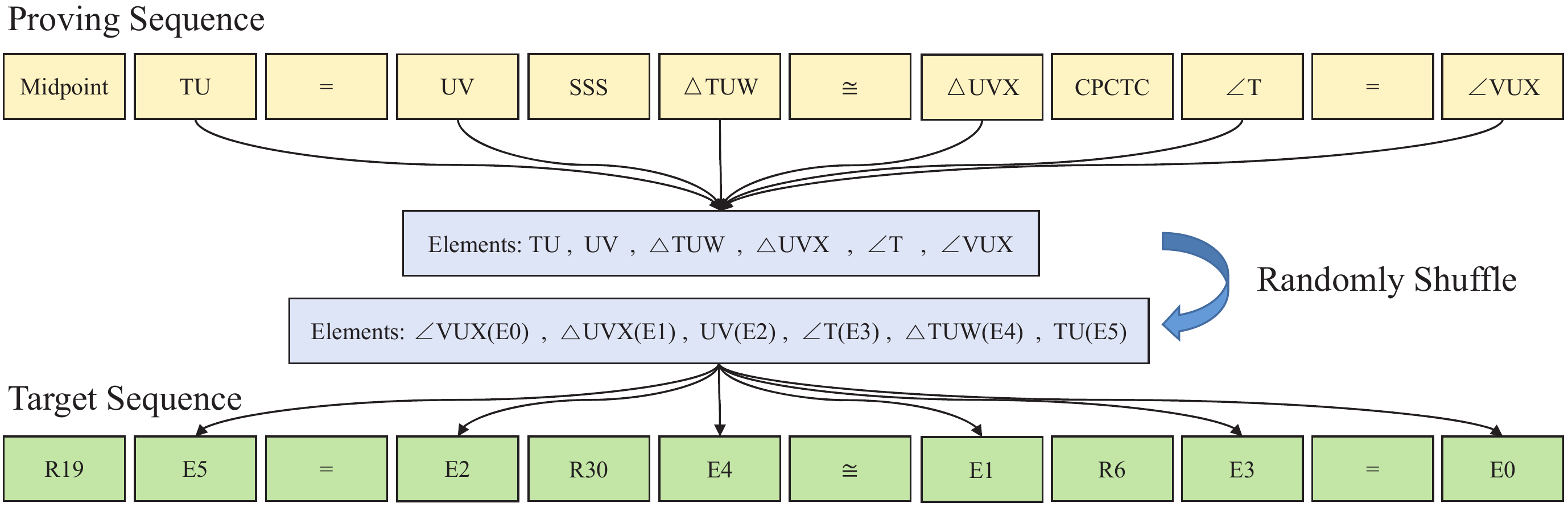}
\end{center}
\vspace{-4mm}
  \caption{
  Illustration of converting proving sequence to the target sequence which is considered as training target for the proving problem.
  }
\vspace{-2mm}
\label{fig:sequence}
\end{figure*}

\subsection{UniGeo Benchmark}

\subsubsection{Data Collection}

We discover an online education website, IXL\footnote{\href{https://www.ixl.com/math/geometry}{https://www.ixl.com/math/geometry}}, which contains various types of geometry problems from high school textbooks. 
We utilize some crawler scripts in Python to crawl a large amount of proving data from this educational website automatically.
After selecting the proving problem carefully, we ask some well-trained workers to check the quality of collected data, such as ensuring that each problem has complete diagram and clear proof.
All the calculation problems are inherited from the GeoQA dataset, containing 4,998 calculation problems with program sequence annotation which corresponds to problem solution and can be predicted by generative models. 
We also organize five well-trained college students to translate the problems in GeoQA dataset from Chinese to English so that the language of the two types of geometry data is consistent.
Finally, we unify these newly collected proving data with the GeoQA dataset and construct our UniGeo benchmark to be a testbed for unified geometry logical reasoning.

\subsubsection{Data Analysis}
\label{data_analysis}

In this section, we mainly analyze the newly collected proving problems in the UniGeo benchmark.
We collect a total of 9,543 proving problems where each data contains a colored geometry diagram, a description text, and the proof with reasons and expressions.
There are totally 37 categories of reasons, which are explanations for each step of the proof, including the reasoning skills or the geometry theorems applied. And the expression is a concrete mathematical proof of each step, consisting of operator and geometry element.
For example, in Figure \ref{fig:introduce}, the reason \textit{Midpoint} represents using the definition of midpoint to get an expression \textit{TU = UV}. \textit{SSS} stands for "side, side, side" and means that we have two congruent triangles with all three sides equal. \textit{CPCTC} stands for "corresponding parts of congruent triangles are congruent", therefore, we get the final expression $\angle T = \angle VUX$.

As shown in Table \ref{table:statistics}, the proving data is divided into train, validation, and test splits with
a ratio of 7:1.5:1.5. The dataset consists of five sub-tasks which also represent five different topics of proving problems: parallel, triangle, quadrangle, congruent, and similarity. The distribution of these types of proving problems can be viewed in Table \ref{table:statistics}. 
In the experiments, we also provide the detailed performance of models on these sub-tasks.

\begin{figure*}[t]
\begin{center}
 \includegraphics[width=1.0\textwidth]{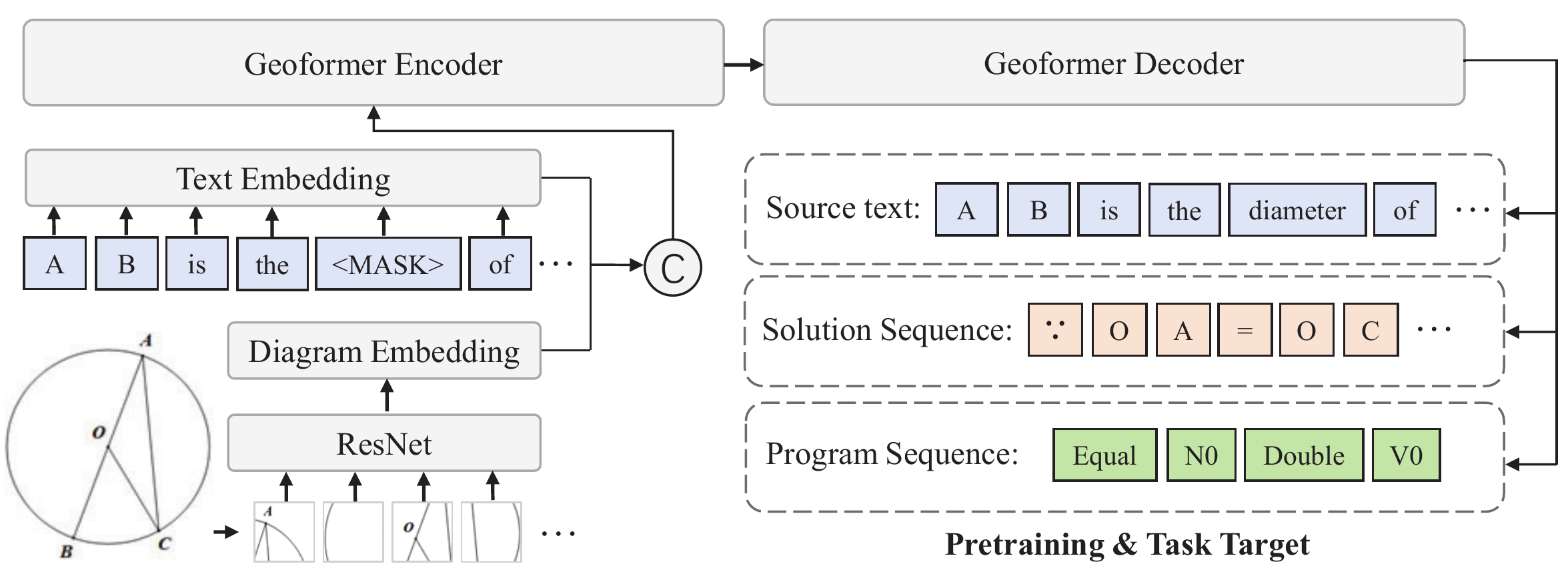}
\end{center}
\vspace{-2mm}
  \caption{
An illustration of our proposed geometric transformer. 
We concatenate the embeddings of text and diagram, which are fed into the transformer encoder-decoder to generate target sequence. 
For pretraining, the targets are source text and the mathematical expressions extracted from the solution. And during the fine-tuning stage, the training objective is program sequence or proving sequence. 
Note that we achieve unified fine-tuning for calculation and proving problems simultaneously. For brevity, this illustration shows only one example of a calculation problem.
  }
\label{fig:method}
\end{figure*}

\begin{table}[t]
    \normalsize
    \centering
    \renewcommand\tabcolsep{5.0pt}
    \begin{tabular}{lc|ccc}
        \toprule	
         & \text{All} & \text{Train}  & \text{Val} & \text{Test}  \\ 
        \midrule	
        \textit{All} & 9,543 & 6,675 & 1,421 & 1,447  \\
        \midrule
        \textit{Parallel} & 443 & 311 & 61 & 71 \\ 
        \textit{Triangle} & 3,035 & 2,134 & 452 & 449 \\
        \textit{Quadrangle} & 1,704 & 1,170 & 260 & 274 \\
        \textit{Congruent} & 2,808 & 1,974 & 414 & 420\\
        \textit{Similarity} & 1,553 & 1,086 & 234 & 233\\
        \bottomrule	
    \end{tabular}
    \vspace{-1mm}
    \caption{Statistics for the proving problems in UniGeo. There are five reasoning sub-tasks for geometry proving. }
\label{table:statistics}
\end{table}

\subsection{Reformulate Expressions in the Proof}

Based on the collected proving data, we aim to reformulate the mathematical expressions as target sequences to unify with the program sequence in calculation problems, thus achieving a reasonable unified geometry reasoning task.
The reasons and expressions in the collected proof are still textual, thus we first translate them into a sequence format.
As shown in Figure \ref{fig:sequence}, we organize the proof as the proving sequence which contains three types of tokens: reasons $R$ (e.g., \textit{Midpoint, SSS, CPCTC}), operators $OP$ (e.g., $=, \cong$), and geometry elements $E$ (e.g., \textit{TU, $\triangle TUW$}, etc.).
The reasons are inserted in front of the proof expressions (including operators and elements) to form the proving sequence.

Moreover, we reformulate the proving sequence as the final target sequence which can be predicted by generative models.
As mentioned in Section \ref{data_analysis}, we have summarized all the reasons into a set, thus, each reason can be considered as a token $R_i$, where $i$ is the index in the predefined reasons set. For the operators, we just reserve their original representation as the tokens.
As for geometry elements, however, we first fetch all the geometry elements in the proving sequence, and construct the list of geometry elements.
To increase the diversity of proving problems, we randomly shuffle these elements to form a new elements list and convert each element in the proving sequence to a token  $E_i$, where $i$ corresponds to its position in the shuffled geometry elements list.
Benefiting from this, we produce diverse target sequences. Even if similar topics may exist in the training and testing sets, the target sequence tends to be completely different, avoiding that the model simply learns some typical proof patterns.
Note that the shuffled elements list will also be added as text to the end of the problem text and fed into the model while training.

In summary, by reformulating the expression-based proof as the target sequence, we define a multimodal high-level reasoning task.
This scheme is adopted for the following reasons.
First, it simplifies the task representation with a clear sequence prediction. 
Second, although the target token is reduced to a smaller space, it is still challenging for models to understand the correspondence between inputs (including problem diagram, text, and the elements to be selected) and target token.
Third, by applying the expression reformulation, we unify the proving problem with the calculation problem to construct the UniGeo benchmark, which requires multiple reasoning capabilities.

\section{Unified Geometric Transformer}

\subsection{Overview}

Although the NGS model \cite{chen2021geoqa} is designed for geometry calculation problems, its performance degrades about 5\% after unified training on the UniGeo (Table \ref{table:result}).
Therefore, based on the VL-T5 \cite{cho2021unifying} model which is capable of handling multiple multimodal tasks uniformly, we propose a geometric transformer (Geoformer) that can conduct comprehensive reasoning on both calculation and proving problems.
Figure \ref{fig:method} demonstrates the structure of the model, which consists of a bidirectional multimodal encoder and an autoregressive text decoder.
In order to promote the performance of Geoformer, we first pretrain it using the solution provided in calculation problems, as well as applying masked LM task to enhance the representation of the text encoder.
At the fine-tuning stage, we train the model end-to-end with calculation and proving problems simultaneously to acquire stronger comprehensive reasoning ability on geometry problem solving, rather than optimizing the model on two tasks separately.

\subsection{Unified Pretraining}

\subsubsection{Mathematical Expression Pretraining}

Different from the popular pretraining paradigm that fine-tuning models with large-scale natural corpus, geometry problems are mainly described by some mathematical languages and also solved by mathematical knowledge, which is far from natural language.
Therefore, we propose Mathematical Expression Pretraining (MEP) to pretrain our unified Geoformer with the mathematical corpus.

\paragraph{Formulating Solution Sequence}

The GeoQA dataset provides a problem solution that explains the idea and process of solving the problem in the form of text description.
Similar to formulating the proof expression into the sequence, mathematical expression in the solution can also be reformulated into solution sequence for prediction.
We remove the redundant text description in the solution and only utilize the mathematical expressions for pretraining. Specifically, we only reserve the geometry elements entities, operation symbols, and numbers, all of which involve abundant geometry mathematical knowledge and can be organized as solution sequence.
In addition, the number in solution is replaced with a token ${NS}_{i}$ where $i$ represents the order that the number appears in the solution. 
Different from taking word-level tokenization for natural text description, we adopt char-level tokenization for geometry elements since some geometry elements share common characters with specific geometric meanings.
For example, both line $OC$ and $\angle OCA$ contain points $O$ and $C$, but this relation will disappear if $OC$ and $\angle OCA$ are considered as basic tokens. 
In summary, the formulated solution sequence has rich mathematical knowledge and can be learned by models to enhance the understanding of mathematical reasoning process.

\subsubsection{Masked Language Modeling}
We also explore applying the Masked Language Modeling (MLM) task for solving geometry problems.
Following \cite{cho2021unifying},  we mask 30\% of input text tokens with \texttt{<mask>} tokens. Then the model is trained to recover the masked text in a unified text generation manner.

\subsection{Fine-tuning Unified Geoformer}

We combine the above two pretraining tasks to pretrain the unified geometric transformer.
After that, fine-tuning the unified Geoformer is straightforward since we have unified the outputs of all downstream tasks into a sequence format.
We load the weights from the pretrained model and keep the weights of the diagram encoder fixed, following the NGS model \cite{chen2021geoqa}.
Then, we optimize the rest parts of the model end-to-end using a mixture of calculation and proving data.

\subsection{Unified Training Objective}

All of the pre-training and fine-tuning tasks in this work are unified in the form of text generation, thus sharing the same training objective. The generation loss $L_g$ is the negative log-likelihood (NLL) of the target sequence:
{
\setlength\abovedisplayskip{0.45pt}
\setlength\belowdisplayskip{0.55pt}
\begin{equation*}
    \mathcal L_g(\bm \theta)=\frac{1}{L}\sum_{t=1}^L \log P_t(y_t|\bm x, y_1, ..., y_{t-1};\bm \theta),
\end{equation*}
}%
where $\bm \theta$ are the parameters of the entire Geoformer architecture except for the diagram encoder, $\bm x$ is the input of both problem text and the extracted diagram feature, $\bm y_t$ are the target tokens, $\bm P_t$ is the distribution of the next token, $\bm L$ is the length of sequence.

\begin{table*}[t]
\centering
\renewcommand\tabcolsep{2.5pt}
\renewcommand{\arraystretch}{0.91}
\resizebox{1.0\linewidth}{!} {
\begin{tabular}{{l}*{12}{c}}
    \toprule
    \multirow{4}{*}{} & \multirow{4}{*}{} &
    \multicolumn{3}{c}{\textbf{Calculation (\%)}} & \multicolumn{6}{c}{\textbf{Proving (\%)}} 	\\
    \cmidrule(lr){3-5} \cmidrule{6-11} 
    \textbf{Methods} & \textbf{Data}  & All & Angle & Length 
    & All & Par. & Tri. & Qua. & Con. & Sim.   \\
    \midrule
    FiLM \cite{perez2017film}  & Calculation  & 31.7 & 34.0 & 29.7  & - & - & - & - & - & -  \\
    RN \cite{santoro2017simple}  &Calculation & 38.0 & 42.8 & 32.5  & - & - & - & - & - & - \\
    MCAN \cite{yu2019deep} &Calculation & 39.7 & 45.0 & 34.6 & - & - & - & - & - & - \\
    BERT \cite{devlin2018bert} & Calculation  &  54.7 & 65.8 & 42.1  &   - & - & - & -  & - & - \\
    NGS \cite{chen2021geoqa} & Calculation  & 56.9 & 69.8 & 39.2  & - & - & - & - & - & - \\
    Geoformer (Ours) & Calculation & 60.3 & 71.5 & \textbf{49.1}  & - & - & - & - & - & - \\
  
    \midrule
    BERT & Proving  &  -& - & - &  48.0 & 15.5 & 48.1 & 28.5 & 49.5 & 77.6 &  \\
    NGS & Proving  & -& - & -  & 53.2 & 13.2 & 56.6 & 29.8 & 57.1 & \textbf{79.4} &  \\
    Geoformer (Ours) & Proving & -& - & - & 55.7 & 19.4 & 68.3 & 20.4 & 60.6 & 72.5\\
    
    \midrule
    BERT & UniGeo  & 52.0 & 63.1 & 39.2  & 48.1 & 15.4 & 48.0 & \textbf{31.7} & 49.5 & 75.1 &  \\
    NGS  & UniGeo & 51.9 & 63.6 & 38.8 & 47.4 & 11.2 & 46.9 & 31.3 & 48.3 & 77.6  \\ 
    Geoformer (Ours) & UniGeo & 60.9 & 72.2 & 48.8  & 55.8 & 18.1 & 68.8 & 20.4 & 60.3 & 73.3 \\
    Geoformer + Pretraining (Ours) & UniGeo & \textbf{62.5} & \textbf{75.5} & 48.8  & \textbf{56.4} & \textbf{19.4} & \textbf{69.4} & 20.4 & \textbf{60.3} & 75.0\\
    \bottomrule	
\end{tabular}
}
\vspace{-1mm}
\caption{The accuracy comparison of various methods and baseline models under different data settings. The newly collected proving problems provide five sub-tasks (as defined in Table \ref{table:statistics}) for evaluation. 
}
\label{table:result}
\end{table*}

\section{Experiments}

\subsection{Experimental Settings}

\paragraph{Datasets}
We conduct experiments on the UniGeo, containing GeoQA \cite{chen2021geoqa} dataset and our newly collected proving problems. The GeoQA dataset involves 4,998 calculation problems with corresponding annotated program sequence, which illustrates the calculating process of the given problems and is considered as training and testing target. Besides, the GeoQA also provides the problem solution which is not utilized by previous works but is used for pretraining in this work.
We also construct a proving dataset with 9,543 problems, which are split to train, validate, and test subsets in a ratio of 7.0: 1.5: 1.5. We further define five sub-tasks: Parallel, Triangle, Quadrangle, Congruent, and Similarity, to provide the detailed performance of models.
To unify geometry reasoning, we also translate the Chinese calculation problems into English, so that the language of calculation and proving problems are consistent.
We also have considered the Inter-GPS dataset \cite{lu2021inter}. However, it mainly adopts the rule-based parser to translate the problem text into formal language and doesn't have the sequence annotation which can be unified with the proving sequence in our work. Therefore, the Inter-GPS dataset is not compatible with unified training on both calculation and proving problems, and we mainly conduct experiments on GeoQA and newly collected proving data.

\paragraph{Evaluation Metrics}
For the calculation problems, we follow the evaluation metrics in GeoQA, i,e, the accuracy of solving all the problems and two main subsets: angle and length problems. 
Following the IsarStep \cite{li2020isarstep}, we adopt top-1 accuracy and top-10 accuracy for evaluating the proofs. 
Top-K accuracy computes the percentage where the ground-truth proof is among the top K generated proving sequence.
Since the models possibly generate alternative valid proving sequences that are not completely consistent with the provided proof, we mainly use more reasonable top-10 accuracy for evaluating the proving problems.

\paragraph{Implementation Details}
We fill the diagram with a white background to make it equal in length and width, and resize it to 224$\times$224, which is further split into 49 patches with a size of 32$\times$32 each.
Then we apply ResNet \cite{he2016deep} to extract patch features which are further mapped into flattened 1D sequences to construct final diagram embeddings. 
Our Geoformer is implemented by PyTorch \cite{paszke2017automatic}. We use the Adam \cite{loshchilov2017decoupled} optimizer with $\beta_1=0.9$ and $\beta_2=0.999$.
The learning rate is $2e^{-4}$, the batch size is set to 10, and models are trained within 100 epochs.
We train our unified Geoformer on randomly shuffled calculation problems and proving problems simultaneously.
For pretraining, we maintain the settings as mentioned above, but replace the training label with the solution sequence and set the learning rate to $5e^{-4}$.

\subsection{Experimental Results}
Table \ref{table:result} demonstrates the results of our methods and baselines on the calculation and proving problems. We divided the experiments into three parts according to the data used by the model, i.e., the calculation problems from the GeoQA dataset, our newly collected proving problems, and the unified benchmark of both calculation and proving problems. A detailed analysis is shown below.

\paragraph{Baselines}
FiLM \cite{perez2017film}, RN \cite{santoro2017simple}, MCAN \cite{yu2019deep} are three multimodal models with strong cross-modal reasoning abilities that well address the compositional language and elementary visual reasoning benchmark, CLEVR \cite{johnson2017clevr}. They can predict the possibly correct option in calculation problems by using visual question answering. However, this approach does not work well in geometry problem solving since the MCAN achieves the answer accuracy of only 39.7\%.
The ``BERT" model here refers to "BERT2Prog + Diagram" in GeoQA that BERT and ResNet are utilized to encode text and diagram data separately. Finally, the features of these two modalities are fused to guide the generation of target sequence.
The NGS model is specially designed for solving the calculation problems in the GeoQA dataset.
We also re-run the experiment on the English version of the GeoQA dataset using the NGS model and obtain a performance of 56.9\%.

\begin{table}[t]
    \renewcommand\tabcolsep{5.0pt}
    \resizebox{1.0\columnwidth}{!} {
    \begin{tabular}{lc|ccc}
        \toprule	
          \text{Methods} & \text{Data}  & \text{Top-1}  & \text{Top-10}   \\ 
        \midrule	
        NGS &  UniGeo  & 17.4 & 47.4  \\
        NGS + Pretraining & UniGeo & 19.2  & 49.6  \\
        Geoformer & UniGeo & 50.2 & 55.8 \\
        Geoformer + Pretraining & UniGeo & \textbf{51.3} & \textbf{56.4} \\
        
        \bottomrule	
    \end{tabular}
    }
    \vspace{-1mm}
    \caption{Performance comparison on proving problems with different evaluation metrics.}
\label{table:metric}
\end{table}

\begin{figure*}[th!]
\begin{center}
 \includegraphics[width=0.96\textwidth]{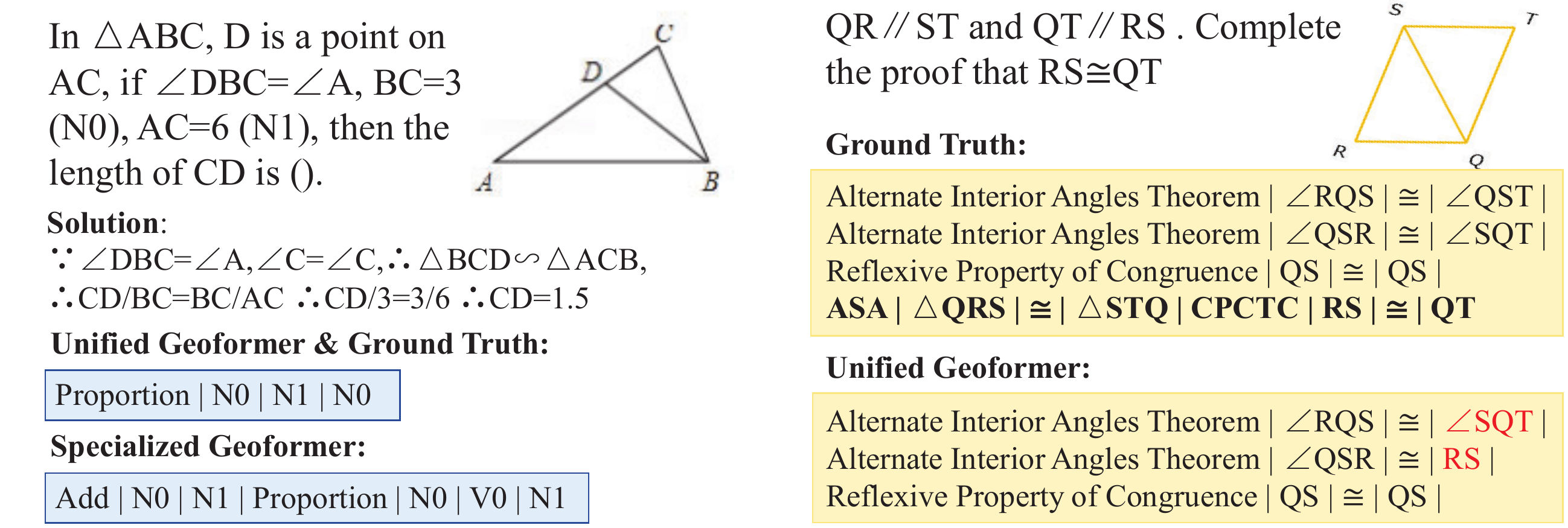}
 \vspace{-2mm}
\end{center}
  \caption{
  The left calculation case shows a situation where a unified Geoformer works better than a task-specialized Geoformer since the related similar triangle knowledge also exists in proving problems.
  Through multi-task learning, the model is enhanced on the understanding of similar triangle problems.
  In the failure proving case on the right, the Geoformer model outputs some incorrect proof (\textcolor{red}{red}) and misses part of the proof (the \textbf{bold} in ground truth).
  }
\label{fig:case}
\end{figure*}

\paragraph{The performance comparison on proving problems}

We conduct some experiments on the collected proving problems.
In Table \ref{table:result}, Par, Tri, Qua, Con, Sim represent five sub-tasks respectively.
When using proving data only, the NGS model achieves a total performance of 53.2\%. 
The proposed Geoformer obtains a top-10 accuracy of 55.7\% on proving problems. 
There is a huge difference in the performance of sub-tasks due to the difficulty of various geometric reasoning skills varies greatly. The accuracy rate of proving parallel related problems is only 19.4\%, while proving similarity is relatively simple, which can reach an accuracy of 72.5\%.
Table \ref{table:metric} also provides the results of top-1 accuracy metric. When applying pretraining on the unified NGS and Geoformer models, we can get a 19.2\% and 51.3\% top-1 accuracy respectively.

\paragraph{The performance of unified training}

Our motivation is to unify the geometry logical reasoning and we have already unified the data format. Thus, apart from training on calculation and proving problems separately, we design the unified Geoformer, which is trained with the mixture of both types of problems.
It can be observed that the NGS model suffers a severe performance decline when trained on both tasks simultaneously, in which the accuracy of calculation and proving problems decrease 5.0\% and 5.8\% respectively. 
However, our proposed Geoformer avoids this phenomenon and obtains an impressive performance on two tasks simultaneously. Specifically, the unified Geoformer achieves 60.9\% and 55.8\% accuracy on calculation and proving problems, outperforming two task-specific Geoformer models on two geometry tasks.
The reasoning ability can be enhanced on both two tasks with the unified formulation.

\paragraph{The effectiveness of pretraining}
To further promote the performance of unified Geoformer, We extract a large number of mathematical expressions from the solution of calculation problem as the pre-training target.
These expressions are rich in implicit math knowledge and can also be formulated as solution sequence.
Applying the pretraining method, the Geoformer+Pretraining model is further improved to 62.5\% and 56.4\% accuracy on calculation and proving problems respectively, obtaining 5.6\% and 3.2\% accuracy improvement compared to task-specialized NGS models, which achieves state-of-the-art performance on UniGeo benchmark.

\begin{table}[t!]
    \renewcommand\tabcolsep{5.0pt}
    \resizebox{1.0\columnwidth}{!} {
    \begin{tabular}{l|ccc}
        \toprule	
          \text{Methods}  & \text{Calculation}  & \text{Proving}   \\ 
        \midrule	
        Geoformer  & 60.9 & 55.8 \\
        Geoformer + MLM & 61.3 & 56.2 \\
        Geoformer + MEP & 61.8 & 56.1 \\
        Geoformer + MLM + MEP & 
        \textbf{62.5} & \textbf{56.4} \\
        \bottomrule	
    \end{tabular}
    }
    \vspace{-1mm}
    \caption{Ablation study for different pretraining methods. MLM and MEP represent masked language modeling and mathematical expression pretraining.
    }
\label{table:ablation}
\end{table}
\subsection{Ablation Study}
We explore the effectiveness of different pretraining settings for the ablation study. In Table \ref{table:ablation}, we experiment unified Geoformer with two pretraining ways: masked language modeling (MLM) and mathematical expression pretraining (MEP).
Using only MLM, the performance of the Geoformer model does not change significantly.
When MEP is used alone, the performance of the model on calculation problems is improved obviously. When using both pre-training methods, the model makes an improvement significantly on both types of problems, obtaining the highest 62.5\% and 56.4\% on calculation and proving problems, respectively. Thus, we apply this setting to the training of Geoformer.

\subsection{Case Study}

As shown in Figure \ref{fig:case}, we conduct a case study for discussing the ability and limitation of our proposed unified Geoformer. 
For the left case, the unified Geoformer works well on the calculation problem, benefiting from the multi-task learning framework. As we can see in the problem solution, it first utilizes the knowledge of similar triangles to get $\triangle  BCD \sim \triangle ACB$, and then lists a proportional relation to get $CD=3/6*3=1.5$.
Compared to task-specialized Geoformer that predicts a wrong program sequence, the prediction made by our unified Geoformer is completely consistent with ground truth. This is probably because the unified model acquires a stronger understanding of similar triangle knowledge after simultaneously training on proving problems (containing many problems proving similar triangles). Therefore, multi-task learning is beneficial in geometry reasoning.
We also select a typical failure case. The unified Geoformer chooses two wrong geometry elements for the proof steps and also fails to give the last two critical proof steps. The geometry problems are still challenging for current neural-based approaches.

\section{Conclusion}
Recently, geometry problem solving has attracted much attention in AI research while previous works mainly focus on geometry calculation problems.
It is significant to explore the unified reasoning abilities of neural models on multiple math tasks.
Therefore, we integrate geometry calculation and proving problems, and construct a unified geometry benchmark, UniGeo, containing 9,543 proving problems with proof reasons and mathematical expressions that can be reformulated as proving sequence to unify with the program sequence of calculation problems.
We also propose a unified Geoformer that can address calculation and proving problems simultaneously.
Besides, a mathematical expression pretraining way is proposed to promote the performance of the unified Geoformer. 
Experiments show that our Geoformer can well address two challenging geometry tasks with a single set of model weights, outperforming task-specialized models and obtaining state-of-the-art performance.


\section*{Limitations}

To explore the logical reasoning ability of neural network models in the geometry domain, we propose a unified method for two major and similar tasks (calculation and proving) in geometry problems. Although we have achieved state-of-the-art performance on these two tasks simultaneously, the unified Geoformer still has some limitations.
First, the answer accuracy of the neural-network-based approaches is still far from the real-world application when addressing such complex tasks which require high-level reasoning ability. 
Second, the data construction of such mathematical logical reasoning tasks requires a heavy manual collection and annotation process, which also limits the type and difficulty of geometry problems, thereby leading to the failure of neural network models to learn and process more sophisticated cases.

\section*{Acknowledgements} 
This work was supported in part by National Key R\&D Program of China under Grant No.2020AAA0109700, 
National Natural Science Foundation of China (NSFC) under Grant No.U19A2073, Grant No.61976233 and Grant No.62206314, 
Guangdong Province Basic and Applied Basic Research (Regional Joint Fund-Key) Grant No.2019B1515120039, 
Guangdong Outstanding Youth Fund (Grant No.2021B1515020061), 
GuangDong Basic and Applied Basic Research Foundation under Grant No.2022A1515011835, 
China Postdoctoral Science Foundation under Grant No.2021M703687, 
Shenzhen Fundamental Research Program (Project No.RCYX20200714114642083)  and CAAI-Huawei MindSpore Open Fund. 
And the Open Project of Anhui Provincial Key Laboratory of Multimodal Cognitive Computation, Anhui University, No.MMC202107. 
We thank MindSpore for the partial support of this work, which is a new deep learning computing framwork\footnote{\href{https://www.mindspore.cn/}{https://www.mindspore.cn/}}.

\bibliographystyle{acl_natbib}
\bibliography{custom}

\begin{thebibliography}{41}
\expandafter\ifx\csname natexlab\endcsname\relax\def\natexlab#1{#1}\fi

\bibitem[{Alayrac et~al.(2022)Alayrac, Donahue, Luc, Miech, Barr, Hasson, Lenc,
  Mensch, Millican, Reynolds et~al.}]{alayrac2022flamingo}
Jean-Baptiste Alayrac, Jeff Donahue, Pauline Luc, Antoine Miech, Iain Barr,
  Yana Hasson, Karel Lenc, Arthur Mensch, Katie Millican, Malcolm Reynolds,
  et~al. 2022.
\newblock Flamingo: a visual language model for few-shot learning.
\newblock \emph{arXiv preprint arXiv:2204.14198}.

\bibitem[{Alvin et~al.(2017)Alvin, Gulwani, Majumdar, and
  Mukhopadhyay}]{alvin2017synthesis}
Chris Alvin, Sumit Gulwani, Rupak Majumdar, and Supratik Mukhopadhyay. 2017.
\newblock Synthesis of solutions for shaded area geometry problems.
\newblock In \emph{The Thirtieth International Flairs Conference}.

\bibitem[{Cao and Xiao(2022)}]{cao2022augmented}
Jie Cao and Jing Xiao. 2022.
\newblock An augmented benchmark dataset for geometric question answering
  through dual parallel text encoding.
\newblock In \emph{Proceedings of the 29th International Conference on
  Computational Linguistics (COLING)}, pages 1511--1520.

\bibitem[{Chen et~al.(2021)Chen, Tang, Qin, Liang, Liu, Xing, and
  Lin}]{chen2021geoqa}
Jiaqi Chen, Jianheng Tang, Jinghui Qin, Xiaodan Liang, Lingbo Liu, Eric~P Xing,
  and Liang Lin. 2021.
\newblock Geoqa: A geometric question answering benchmark towards multimodal
  numerical reasoning.
\newblock In \emph{Findings of the Association for Computational Linguistics
  (ACL-IJCNLP)}.

\bibitem[{Cho et~al.(2021)Cho, Lei, Tan, and Bansal}]{cho2021unifying}
Jaemin Cho, Jie Lei, Hao Tan, and Mohit Bansal. 2021.
\newblock Unifying vision-and-language tasks via text generation.
\newblock In \emph{International Conference on Machine Learning}, pages
  1931--1942. PMLR.

\bibitem[{Chou et~al.(1996)Chou, Gao, and Zhang}]{chou1996automated}
Shang-Ching Chou, Xiao-Shan Gao, and Jing-Zhong Zhang. 1996.
\newblock Automated generation of readable proofs with geometric invariants.
\newblock \emph{Journal of Automated Reasoning}, 17(3):325--347.

\bibitem[{Chou et~al.(2000)Chou, Gao, and Zhang}]{chou2000deductive}
Shang-Ching Chou, Xiao-Shan Gao, and Jing-Zhong Zhang. 2000.
\newblock A deductive database approach to automated geometry theorem proving
  and discovering.
\newblock \emph{Journal of Automated Reasoning}, 25(3):219--246.

\bibitem[{Devlin et~al.(2018)Devlin, Chang, Lee, and
  Toutanova}]{devlin2018bert}
Jacob Devlin, Ming-Wei Chang, Kenton Lee, and Kristina Toutanova. 2018.
\newblock Bert: Pre-training of deep bidirectional transformers for language
  understanding.
\newblock \emph{arXiv preprint arXiv:1810.04805}.

\bibitem[{Gan et~al.(2019)Gan, Yu, Zhang, and Wang}]{gan2019automatically}
Wenbin Gan, Xinguo Yu, Ting Zhang, and Mingshu Wang. 2019.
\newblock Automatically proving plane geometry theorems stated by text and
  diagram.
\newblock \emph{International Journal of Pattern Recognition and Artificial
  Intelligence}, 33(07):1940003.

\bibitem[{Gelernter et~al.(1960)Gelernter, Hansen, and
  Loveland}]{gelernter1960empirical}
Herbert Gelernter, James~R Hansen, and Donald~W Loveland. 1960.
\newblock Empirical explorations of the geometry theorem machine.
\newblock In \emph{Western Joint IRE-AIEE-ACM Computer Conference}, pages
  143--149.

\bibitem[{He et~al.(2016)He, Zhang, Ren, and Sun}]{he2016deep}
Kaiming He, Xiangyu Zhang, Shaoqing Ren, and Jian Sun. 2016.
\newblock Deep residual learning for image recognition.
\newblock In \emph{Proceedings of the IEEE conference on computer vision and
  pattern recognition}, pages 770--778.

\bibitem[{Johnson et~al.(2017)Johnson, Hariharan, Van Der~Maaten, Fei-Fei,
  Lawrence~Zitnick, and Girshick}]{johnson2017clevr}
Justin Johnson, Bharath Hariharan, Laurens Van Der~Maaten, Li~Fei-Fei,
  C~Lawrence~Zitnick, and Ross Girshick. 2017.
\newblock Clevr: A diagnostic dataset for compositional language and elementary
  visual reasoning.
\newblock In \emph{Proceedings of the IEEE Conference on Computer Vision and
  Pattern Recognition (CVPR)}, pages 2901--2910.

\bibitem[{Li et~al.(2022)Li, Li, Xiong, and Hoi}]{li2022blip}
Junnan Li, Dongxu Li, Caiming Xiong, and Steven Hoi. 2022.
\newblock Blip: Bootstrapping language-image pre-training for unified
  vision-language understanding and generation.
\newblock In \emph{ICML}.

\bibitem[{Li et~al.(2020)Li, Yu, Wu, and Paulson}]{li2020isarstep}
Wenda Li, Lei Yu, Yuhuai Wu, and Lawrence~C Paulson. 2020.
\newblock Isarstep: a benchmark for high-level mathematical reasoning.
\newblock In \emph{The International Conference on Learning Representations
  (ICLR)}.

\bibitem[{Loshchilov and Hutter(2017)}]{loshchilov2017decoupled}
Ilya Loshchilov and Frank Hutter. 2017.
\newblock Decoupled weight decay regularization.
\newblock \emph{arXiv preprint arXiv:1711.05101}.

\bibitem[{Lu et~al.(2021{\natexlab{a}})Lu, Gong, Jiang, Qiu, Huang, Liang, and
  Zhu}]{lu2021inter}
Pan Lu, Ran Gong, Shibiao Jiang, Liang Qiu, Siyuan Huang, Xiaodan Liang, and
  Song-Chun Zhu. 2021{\natexlab{a}}.
\newblock Inter-gps: Interpretable geometry problem solving with formal
  language and symbolic reasoning.
\newblock In \emph{The Joint Conference of the 59th Annual Meeting of the
  Association for Computational Linguistics and the 11th International Joint
  Conference on Natural Language Processing (ACL-IJCNLP)}.

\bibitem[{Lu et~al.(2022)Lu, Qiu, Chang, Wu, Zhu, Rajpurohit, Clark, and
  Kalyan}]{lu2022dynamic}
Pan Lu, Liang Qiu, Kai-Wei Chang, Ying~Nian Wu, Song-Chun Zhu, Tanmay
  Rajpurohit, Peter Clark, and Ashwin Kalyan. 2022.
\newblock Dynamic prompt learning via policy gradient for semi-structured
  mathematical reasoning.
\newblock \emph{arXiv preprint arXiv:2209.14610}.

\bibitem[{Lu et~al.(2021{\natexlab{b}})Lu, Qiu, Chen, Xia, Zhao, Zhang, Yu,
  Liang, and Zhu}]{lu2021iconqa}
Pan Lu, Liang Qiu, Jiaqi Chen, Tony Xia, Yizhou Zhao, Wei Zhang, Zhou Yu,
  Xiaodan Liang, and Song-Chun Zhu. 2021{\natexlab{b}}.
\newblock Iconqa: A new benchmark for abstract diagram understanding and visual
  language reasoning.
\newblock In \emph{The 35th Conference on Neural Information Processing Systems
  Track on Datasets and Benchmarks (NeurIPS)}.

\bibitem[{Mishra et~al.(2022{\natexlab{a}})Mishra, Finlayson, Lu, Tang,
  Welleck, Baral, Rajpurohit, Tafjord, Sabharwal, Clark, and
  Kalyan}]{mishra2022lila}
Swaroop Mishra, Matthew Finlayson, Pan Lu, Leonard Tang, Sean Welleck, Chitta
  Baral, Tanmay Rajpurohit, Oyvind Tafjord, Ashish Sabharwal, Peter Clark, and
  Ashwin Kalyan. 2022{\natexlab{a}}.
\newblock Lila: A unified benchmark for mathematical reasoning.
\newblock In \emph{The 2022 Conference on Empirical Methods in Natural Language
  Processing (EMNLP)}.

\bibitem[{Mishra et~al.(2022{\natexlab{b}})Mishra, Mitra, Varshney, Sachdeva,
  Clark, Baral, and Kalyan}]{mishra2022numglue}
Swaroop Mishra, Arindam Mitra, Neeraj Varshney, Bhavdeep Sachdeva, Peter Clark,
  Chitta Baral, and Ashwin Kalyan. 2022{\natexlab{b}}.
\newblock Numglue: A suite of fundamental yet challenging mathematical
  reasoning tasks.
\newblock In \emph{Proceedings of the 60th Annual Meeting of the Association
  for Computational Linguistics (ACL)}, pages 3505--3523.

\bibitem[{Paszke et~al.(2017)Paszke, Gross, Chintala, Chanan, Yang, DeVito,
  Lin, Desmaison, Antiga, and Lerer}]{paszke2017automatic}
Adam Paszke, Sam Gross, Soumith Chintala, Gregory Chanan, Edward Yang, Zachary
  DeVito, Zeming Lin, Alban Desmaison, Luca Antiga, and Adam Lerer. 2017.
\newblock Automatic differentiation in pytorch.
\newblock \emph{.}

\bibitem[{Perez et~al.(2017)Perez, Strub, De~Vries, Dumoulin, and
  Courville}]{perez2017film}
Ethan Perez, Florian Strub, Harm De~Vries, Vincent Dumoulin, and Aaron
  Courville. 2017.
\newblock Film: Visual reasoning with a general conditioning layer.
\newblock \emph{arXiv preprint arXiv:1709.07871}.

\bibitem[{Qin et~al.(2021)Qin, Liang, Hong, Tang, and Lin}]{NS-Solver}
Jinghui Qin, Xiaodan Liang, Yining Hong, Jianheng Tang, and Liang Lin. 2021.
\newblock Neural-symbolic solver for math word problems with auxiliary tasks.
\newblock In \emph{Proceedings of the 59th Annual Meeting of the Association
  for Computational Linguistics and the 11th International Joint Conference on
  Natural Language Processing (ACL-IJCNLP)}.

\bibitem[{Qin et~al.(2020)Qin, Lin, Liang, Zhang, and Lin}]{sau-solver}
Jinghui Qin, Lihui Lin, Xiaodan Liang, Rumin Zhang, and Liang Lin. 2020.
\newblock Semantically-aligned universal tree-structured solver for math word
  problems.
\newblock In \emph{Proceedings of the 2020 Conference on Empirical Methods in
  Natural Language Processing (EMNLP)}, pages 3780--3789.

\bibitem[{Raffel et~al.(2020)Raffel, Shazeer, Roberts, Lee, Narang, Matena,
  Zhou, Li, Liu et~al.}]{raffel2020exploring}
Colin Raffel, Noam Shazeer, Adam Roberts, Katherine Lee, Sharan Narang, Michael
  Matena, Yanqi Zhou, Wei Li, Peter~J Liu, et~al. 2020.
\newblock Exploring the limits of transfer learning with a unified text-to-text
  transformer.
\newblock \emph{J. Mach. Learn. Res.}, 21(140):1--67.

\bibitem[{Sachan et~al.(2020)Sachan, Dubey, Hovy, Mitchell, Roth, and
  Xing}]{sachan2020discourse}
Mrinmaya Sachan, Avinava Dubey, Eduard~H Hovy, Tom~M Mitchell, Dan Roth, and
  Eric~P Xing. 2020.
\newblock Discourse in multimedia: A case study in extracting geometry
  knowledge from textbooks.
\newblock \emph{Computational Linguistics}, 45(4):627--665.

\bibitem[{Sachan et~al.(2017)Sachan, Dubey, and Xing}]{sachan2017textbooks}
Mrinmaya Sachan, Kumar Dubey, and Eric Xing. 2017.
\newblock From textbooks to knowledge: A case study in harvesting axiomatic
  knowledge from textbooks to solve geometry problems.
\newblock In \emph{Proceedings of Empirical Methods in Natural Language
  Processing (EMNLP)}, pages 773--784.

\bibitem[{Sachan and Xing(2017)}]{sachan2017learning}
Mrinmaya Sachan and Eric Xing. 2017.
\newblock Learning to solve geometry problems from natural language
  demonstrations in textbooks.
\newblock In \emph{Proceedings of the 6th Joint Conference on Lexical and
  Computational Semantics}, pages 251--261.

\bibitem[{Santoro et~al.(2017)Santoro, Raposo, Barrett, Malinowski, Pascanu,
  Battaglia, and Lillicrap}]{santoro2017simple}
Adam Santoro, David Raposo, David~G Barrett, Mateusz Malinowski, Razvan
  Pascanu, Peter Battaglia, and Timothy Lillicrap. 2017.
\newblock A simple neural network module for relational reasoning.
\newblock In \emph{Advances in neural information processing systems
  (NeurIPS)}, pages 4967--4976.

\bibitem[{Seo et~al.(2014)Seo, Hajishirzi, Farhadi, and
  Etzioni}]{seo2014diagram}
Minjoon Seo, Hannaneh Hajishirzi, Ali Farhadi, and Oren Etzioni. 2014.
\newblock Diagram understanding in geometry questions.
\newblock In \emph{Proceedings of the AAAI Conference on Artificial
  Intelligence (AAAI)}.

\bibitem[{Seo et~al.(2015)Seo, Hajishirzi, Farhadi, Etzioni, and
  Malcolm}]{seo2015solving}
Minjoon Seo, Hannaneh Hajishirzi, Ali Farhadi, Oren Etzioni, and Clint Malcolm.
  2015.
\newblock Solving geometry problems: Combining text and diagram interpretation.
\newblock In \emph{Proceedings of Empirical Methods in Natural Language
  Processing (EMNLP)}, pages 1466--1476.

\bibitem[{Tafjord et~al.(2020)Tafjord, Mishra, and
  Clark}]{tafjord2020proofwriter}
Oyvind Tafjord, Bhavana~Dalvi Mishra, and Peter Clark. 2020.
\newblock Proofwriter: Generating implications, proofs, and abductive
  statements over natural language.
\newblock In \emph{Findings of the Association for Computational Linguistics
  (ACL-IJCNLP)}.

\bibitem[{Welleck et~al.(2021)Welleck, Liu, Bras, Hajishirzi, Choi, and
  Cho}]{welleck2021naturalproofs}
Sean Welleck, Jiacheng Liu, Ronan~Le Bras, Hannaneh Hajishirzi, Yejin Choi, and
  Kyunghyun Cho. 2021.
\newblock Naturalproofs: Mathematical theorem proving in natural language.
\newblock In \emph{The 35th Conference on Neural Information Processing Systems
  (NeurIPS)}.

\bibitem[{Yang et~al.(2022{\natexlab{a}})Yang, Qin, Chen, and
  Liang}]{yang-etal-2022-unbiased}
ZhiCheng Yang, Jinghui Qin, Jiaqi Chen, and Xiaodan Liang. 2022{\natexlab{a}}.
\newblock Unbiased math word problems benchmark for mitigating solving bias.
\newblock In \emph{Findings of the Association for Computational Linguistics:
  NAACL 2022}, pages 1401--1408. Association for Computational Linguistics.

\bibitem[{Yang et~al.(2022{\natexlab{b}})Yang, Qin, Chen, Lin, and
  Liang}]{yang2022logicsolver}
Zhicheng Yang, Jinghui Qin, Jiaqi Chen, Liang Lin, and Xiaodan Liang.
  2022{\natexlab{b}}.
\newblock Logicsolver: Towards interpretable math word problem solving with
  logical prompt-enhanced learning.
\newblock \emph{arXiv preprint arXiv:2205.08232}.

\bibitem[{Ye et~al.(2011)Ye, Chou, and Gao}]{jgex2011}
Zheng Ye, Shang-Ching Chou, and Xiao-Shan Gao. 2011.
\newblock An introduction to java geometry expert.
\newblock In \emph{International Workshop on Automated Deduction in Geometry},
  pages 189--195. Springer.

\bibitem[{Yu et~al.(2019{\natexlab{a}})Yu, Wang, Gan, He, and
  Ye}]{yu2019framework}
Xinguo Yu, Mingshu Wang, Wenbin Gan, Bin He, and Nan Ye. 2019{\natexlab{a}}.
\newblock A framework for solving explicit arithmetic word problems and proving
  plane geometry theorems.
\newblock \emph{International Journal of Pattern Recognition and Artificial
  Intelligence}, 33(07):1940005.

\bibitem[{Yu et~al.(2019{\natexlab{b}})Yu, Yu, Cui, Tao, and Tian}]{yu2019deep}
Zhou Yu, Jun Yu, Yuhao Cui, Dacheng Tao, and Qi~Tian. 2019{\natexlab{b}}.
\newblock Deep modular co-attention networks for visual question answering.
\newblock In \emph{Proceedings of the IEEE conference on computer vision and
  pattern recognition (CVPR)}, pages 6281--6290.

\bibitem[{Zhang et~al.(2020{\natexlab{a}})Zhang, LEE, Lim, Qin, Wang, Shao, Sun
  et~al.}]{tsrmd}
Jipeng Zhang, Ka~Wei LEE, Ee-Peng Lim, Wei Qin, Lei Wang, Jie Shao, Qianru Sun,
  et~al. 2020{\natexlab{a}}.
\newblock Teacher-student networks with multiple decoders for solving math word
  problem.
\newblock In \emph{Proceedings of the Twenty-Ninth International Joint
  Conference on Artificial Intelligence (IJCAI)}.

\bibitem[{Zhang et~al.(2020{\natexlab{b}})Zhang, Wang, Lee, Bin, Wang, Shao,
  and Lim}]{graph2tree}
Jipeng Zhang, Lei Wang, Roy Ka-Wei Lee, Yi~Bin, Yan Wang, Jie Shao, and Ee-Peng
  Lim. 2020{\natexlab{b}}.
\newblock Graph-to-tree learning for solving math word problems.
\newblock In \emph{Proceedings of the 58th Annual Meeting of the Association
  for Computational Linguistics (ACL)}, pages 3928--3937.

\bibitem[{Zhang et~al.(2022)Zhang, Yin, Hao, and Liu}]{ijcai2022p228}
Ming-Liang Zhang, Fei Yin, Yi-Han Hao, and Cheng-Lin Liu. 2022.
\newblock Plane geometry diagram parsing.
\newblock In \emph{Proceedings of the Thirty-First International Joint
  Conference on Artificial Intelligence (IJCAI)}, pages 1636--1643.

\end{thebibliography}

\end{document}